%% file: main.tex
\PassOptionsToPackage{table}{xcolor}
\RequirePackage{xcolor}
\documentclass{bmvc2k}
\usepackage[utf8]{inputenc}
\usepackage{times}

\usepackage{amsmath}
\usepackage{amssymb}
\usepackage{booktabs}
\usepackage{etoolbox}
\usepackage{floatrow}
\usepackage{graphicx}
\usepackage{lipsum}
\usepackage{microtype}
\usepackage{multirow}
\usepackage{pifont}
\usepackage{nicefrac}
\usepackage{url}
\usepackage{cleveref}

\usepackage{pgfplots}
\pgfplotsset{compat=1.14}
\definecolor{col1}{RGB}{232, 161, 148}
\definecolor{col2}{RGB}{148, 187, 232}

\newsavebox{\measurebox}

\newcommand{\cmark}{\ding{51}}
\newcommand{\xmark}{\ding{55}}

\newif\ifdark\darktrue
\IfFileExists{/Users/vedaldi/.bash_profile}{
\definecolor{backgroundColor}{gray}{1}
\definecolor{textColor}{gray}{0}
\ifdark
\definecolor{backgroundColor}{gray}{0.15}
\definecolor{textColor}{gray}{.8}
\fi
\hbadness=\maxdimen
\vbadness=\maxdimen
\vfuzz=30pt
\hfuzz=30pt
\pagecolor{backgroundColor}
\color{textColor}
}{}

\title{Calibrating Self-supervised Monocular Depth Estimation}

\addauthor{Robert McCraith}{robert@robots.ox.ac.uk}{1}
\addauthor{Lukas Neumann}{lukas@robots.ox.ac.uk}{1}
\addauthor{Andrea Vedaldi}{vedaldi@robots.ox.ac.uk}{1}

\addinstitution{
 University of Oxford
}
\addinstitution{
 University of Oxford
}
\addinstitution{
 University of Oxford
}

\runninghead{McCraith, Neumann, Vedaldi}{Calibrating Self-supervised Monocular Depth}

\begin{document}

\maketitle

\begin{abstract}
In the recent years, many methods demonstrated the ability of neural networks to learn depth and pose changes in a sequence of images, using only self-supervision as the training signal. Whilst the networks achieve good performance, the often over-looked detail is that due to the inherent ambiguity of monocular vision they predict depth up to a unknown scaling factor. The scaling factor is then typically obtained from the LiDAR ground truth at test time, which severely limits practical applications of these methods.

In this paper, we show that incorporating prior information about the camera configuration and the environment, we can remove the scale ambiguity and predict depth directly, still using the self-supervised formulation and not relying on any additional sensors.
\end{abstract}

\input{intro}
\input{previouswork}
\input{method}
\input{experiments}

\input{conclusion}
\bibliography{egbib}
\end{document}

%% file: intro.tex
\section{Introduction}\label{s:intro}

Depth estimation is an important computer vision problem with applications in robotics, autonomous driving, augmented reality and scene understanding~\cite{poggi2018towards,zhou2017,Fu_2018_CVPR,yang2019inferring,8593864}.
Of particular theoretical and practical interest is estimating depth from a single RGB image, also known as \emph{monocular depth estimation}.
When multiple views are available, depth can be inferred from geometric principles by triangulating image correspondences;
however, when only a single view is available, triangulation is not possible and the problem is ill posed.
Despite this difficulty, reliable and accurate monocular depth estimation is critical for safety in many applications, with autonomous vehicles being a prime example.

Notably, humans are perfectly able to drive a car with just one eye, suggesting that they can infer depth well enough even from a single view.
However, doing so requires prior information on the visual appearance and real-world sizes of typical scene elements, which can then be used to estimate the distance of known objects from the camera.
In machine vision, this prior can be learned from 2D images labelled with ground-truth 3D information, extracted from a different sensing modality such as a LiDAR\@.
When using additional sensors is impractical, self-supervised learning can be used instead~\cite{monodepth17,monodepthv2,zhou2017,mancini2016fast}. In \textit{self-supervised learning}, certain relations or consistency of inputs, rather than ground truth labels, are exploited to train the system.
In depth estimation, one typically uses the consistency between subsequent video frames or between stereo image pairs~\cite{monodepth17,zhou2017}.

Whilst recent self-supervised monocular depth estimation methods achieve impressive performance, approaching fully-supervised systems, they all share an important practical limitation which limits their usefulness in real-world applications.
Since reconstructing 3D geometry from images has an inherent scale ambiguity~\cite{faugeras01the-geometry}, and since self-supervised methods only use visual inputs for training~\cite{monodepth17,monodepthv2}, they do not predict the depth map $d_I$ directly, but rather a scaled version $\Phi(I)$ of it relate to the true depth by an unknown scaling factor $\alpha_I$, in the sense that $d_I = \alpha_I\Phi(I)$.
For evaluation, the scaling factor $\alpha_I$ is not predicted but calculated at \emph{test time from the ground truth}, usually as the the ratio between the median of the predicted depth values and the median of the ground-truth depth values.
Furthermore, a different scaling factor is computed for each test image individually~\cite{monodepth17}.
However, in practical applications ground truth 3D data is not available to calibrate the system, especially in production.
Thus, the problem is how to \emph{calibrate} self-supervised depth estimation in order to obtain a physically-accurate prediction, without requiring the use of additional sensors.
In this paper, we show that, in a driving scenario, this problem can indeed be solved reliably, robustly and efficiently assuming only knowledge of the camera intrinsics and a very limited amount of additional prior information on the geometry of the system.
The output of our technique is a properly calibrated depth map, expressed in meters.
The method is applicable to any self-supervised training paradigm and does not require any additional 3D ground truth at training or testing time.
This is different from previous monocular depth estimation methods which discounting the scale ambiguity at test time, or use additional senors to remove it~\cite{packnet}.
Thus, we make two key contributions in this paper.
First, we bring to the attention of the computer vision community the problem of calibrating self-supervised monocular depth estimation systems without resorting to additional sensors such as LiDARs.
This is of clear importance if we wish these systems to be of direct practical value.
Furthermore, we analyze to what extent the state-of-the-art existing methods depend on the availability of such data.
Second, we propose a simple and yet very efficient calibration technique that \emph{does not} make use of any additional sensors, especially of a complex and expensive nature such as a LiDAR\@.
Instead, our method is `vision closed' at training as well as a test time, in the sense that, just like self-supervised monocular depth estimation methods, it only requires images as input.
The only additional information required for calibration is the approximate knowledge of a single constant which is trivially obtained from the construction of the system.
The rest of the paper is structured as follows. In \cref{sec:previouswork} we give an overview of the state of the art, in \cref{sec:method} our method is presented. Experimental validation is given in \cref{sec:experiments} and the paper is concluded in \cref{sec:conclusion}.

%% file: previouswork.tex
\section{Previous Work}\label{sec:previouswork}
\paragraph{Depth Estimation.}

Because of the scale ambiguity inherent to predicting the depth from a single image, monocular depth estimation is an ill-posed problem and other (prior) knowledge has to be incorporated to remove the ambiguity.
\citet{scharstein2002a} and more recently \citet{FlynnNPS15} use classical  geometry to extract point-to-point matches between images and use triangulation to estimate depth.

\begin{figure}
    \centering
    \input{figs/MD_scaling_factor_LiDAR}
    \caption{Scaling factor inferred from LiDAR ground truth for every image in the KITTI \textbf{test} subset, as used in \citet{monodepthv2}\vspace{-10pt}}\label{fig:monodepthscalingfactor}
    \vspace{-10pt}
\end{figure}
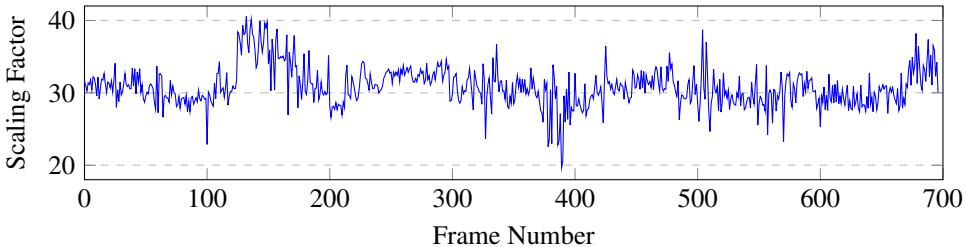

The emergence of deep learning re-formulated the problem as a dense scene segmentation problem, where each image pixel is directly assigned a real value corresponding to the depth. A deep network is then trained to predict depth using supervision either from LiDAR, a RGB-D camera or a stereo pair. \citet{EigenF14} regress depth in multiple scales, refining the depth maps from low  to high spatial resolution. \citet{XieGF16} improve network architecture by adding skip connections, so that the network can also benefit from high resolution information. \citet{LainaRBTN16} use de-convolutional segments to refine depth in a coarse-to-fine manner~\cite{GargBR16,KuznietsovSL17}, while \citet{GargBR16} use CRFs to improve fine details.
DORN~\cite{Fu_2018_CVPR} introduced a novel discretized depth representation, which modifies the typical regression task into a classification problem, and using a novel loss function (ordinal regression) they significantly improve accuracy. More recently, \citet{Lee19} use local planar guidance layers to improve performance.
In practice, there are indeed other modalities/sensors that can be used to get depth --- such as LiDAR, radar or a stereo camera pair. They all however have their own limitations: LiDARs are quite sensitive to weather~\cite{Bijelic_2018}, the depth maps are sparse which may not be enough for far-away or small objects and they have low refresh rate. Similarly, radars suffer from reflections and interference and they struggle to detect small or slowly-moving objects~\cite{radar-shortcomings}.
Stereo is very sensitive to precise calibration and the two cameras can become misaligned over time, greatly reducing the depth map accuracy.
Generalizing, there is a clear need for redundancy --- a moving autonomous vehicle or a robot in urban environment cannot simply stop working if for example one of the cameras becomes occluded by dirt or if the weather is not perfect, as that would be potentially very unsafe. By having a reliable monocular method, this can be used in sensor fusion or as a fallback method, thus improving the overall safety of the system.
\paragraph{Self-Supervised Methods.}
The main struggle for the monocular depth methods is the requirement of vasts amounts of training data. The ground truth is typically captured by LiDARs, but this is expensive especially if large variety of driving scenarios and countries has to be covered, and the output is only a sparse point cloud. To alleviate the requirement of having expensive ground truth, recently there has been a surge in interest in unsupervised methods for depth map prediction. \citet{XieGF16} used stereo images in training discrete values for VR and 3D video applications and \citet{GargBR16} extended this approach to continuous values. More recently, Monodepth~\cite{monodepth17} added a left-right depth consistency and SfMLearner~\cite{zhou2017} generalized the approach to monocular sequences at training time, by predicting pose change between two sequential video frames.
The sequential nature of the data however introduced some new challenges especially for non-stationary objects, which was addressed in Monodepth~V2~\cite{monodepthv2}, that incorporates a loss which automatically excludes pixels which have become occluded or which correspond to moving objects. Similarly, \citet{Casser19} decompose the image into rigid and non-rigid component, thus reducing the re-projection error.
All the above methods~\cite{monodepth17, monodepthv2} however share the same weakness which severely limits their applicability --- the depth estimate is not calibrated, unlike the supervised methods\cite{Fu_2018_CVPR}. In other words, the depth values of self-supervised methods are not in meters but in some arbitrary unit, which moreover \textit{differs frame by frame} (see \cref{fig:monodepthscalingfactor}), and therefore these methods cannot be directly used to reason about the surrounding 3D world.

%% file: figs/MD_scaling_factor_LiDAR.tex
\begin{tikzpicture}
\begin{axis}[
    xlabel={Frame Number},
    ylabel={Scaling Factor},
    xmin=0, xmax=700,
    ymin=18, ymax=42,
    xtick={0,100,200,300,400,500,600,700},
    ytick={20,  30, 40},
    legend pos=north west,
    ymajorgrids=true,
    grid style=dashed,
    width=\textwidth,
    height=0.3\textwidth,
]
\addplot[color=blue]
    coordinates {
   (0,31.65204322540461)(1,30.96776969875401)(2,30.281877843462016)(3,31.36663225413992)(4,31.158820489444093)(5,31.83390597075939)(6,30.413801448595823)(7,32.21420265322362)(8,32.05871089642458)(9,31.123488132251744)(10,29.558259669703236)(11,32.16518330376839)(12,29.156737181157737)(13,30.70281226339073)(14,31.888770383697175)(15,31.772001806825653)(16,32.65762208492342)(17,31.28816534078481)(18,30.033701741439152)(19,31.47467676879569)(20,31.32546669303493)(21,30.653926999649567)(22,31.65153007483098)(23,30.086199474988128)(24,31.74756443311302)(25,34.08505245746636)(26,28.002563284210627)(27,30.429521451531063)(28,28.796059954341185)(29,31.491315750752037)(30,29.7274012768539)(31,30.173419648314393)(32,31.231176204855156)(33,32.66145890648529)(34,31.207369452305354)(35,30.86966893235581)(36,31.102828640036797)(37,30.559191768897392)(38,30.57552668655962)(39,33.42092010920099)(40,30.926013557204698)(41,33.33298617129612)(42,29.869348962588514)(43,32.88547724128489)(44,31.425114797619106)(45,33.33650149550385)(46,30.062715260804996)(47,31.13072150167483)(48,29.169488947572987)(49,28.70209025777417)(50,29.541884064703783)(51,30.090710117573067)(52,29.866682426956544)(53,31.000315650248197)(54,31.251859866654037)(55,30.939184895437474)(56,30.519312956524967)(57,28.700060287172473)(58,27.737602594608784)(59,33.68775445925022)(60,27.252632160103833)(61,32.626687784652326)(62,32.647391477794734)(63,32.27291434425933)(64,26.635354753373505)(65,29.46872384481191)(66,29.653722272562582)(67,29.303905786855626)(68,29.28848781713683)(69,28.726766027131653)(70,31.727684663821748)(71,29.7228409904563)(72,30.930582072878927)(73,30.03540032359509)(74,29.85710650023493)(75,28.108101679630813)(76,28.38069513446423)(77,29.486160317194727)(78,27.88454601563715)(79,29.163225351571313)(80,29.275496190395643)(81,29.362915556432775)(82,28.424961891190552)(83,29.64360650285629)(84,27.39139088196807)(85,28.883515364232085)(86,27.408527448677376)(87,29.375092494226244)(88,29.497218444067766)(89,28.438871649926448)(90,28.890127212610125)(91,29.32272674930033)(92,28.766778651737337)(93,30.360526010767973)(94,29.16011782397032)(95,28.915735815750097)(96,29.63709553852976)(97,29.514636939099166)(98,29.12188726515017)(99,29.233525987697455)(100,22.8731321953116)(101,30.14752075342038)(102,29.776735493398085)(103,28.91908154191511)(104,28.5330955172221)(105,28.16927323561872)(106,31.36280143791885)(107,29.20000249220513)(108,32.67161536353662)(109,32.685961122069415)(110,34.30184285427516)(111,30.08557187948968)(112,30.447223739943045)(113,30.703372500508287)(114,29.37827681579532)(115,30.66001632402783)(116,32.79026673110834)(117,28.303911732049034)(118,29.55207817609037)(119,30.165861504743933)(120,30.16434904557277)(121,30.928628207844078)(122,30.59042524131282)(123,30.595918447041093)(124,31.681121424849625)(125,38.38490151211655)(126,37.982661206905526)(127,36.880250301021974)(128,35.202618988331665)(129,39.34485663493974)(130,38.06929436214763)(131,38.23704063128402)(132,40.553361048964234)(133,36.83936575667607)(134,35.54706093487007)(135,38.7712238718566)(136,40.17861451272353)(137,38.090912510884365)(138,37.70997140729051)(139,35.974501104384856)(140,33.47050999880969)(141,34.0725559841587)(142,36.72852001901867)(143,39.95885288739862)(144,37.09573773526631)(145,35.612848156391934)(146,36.88808939467799)(147,39.61685199994977)(148,38.95574991703813)(149,39.76153688732912)(150,33.97860365304046)(151,35.910435258219074)(152,30.28513020935118)(153,37.414874552678064)(154,35.37406387556379)(155,30.376602532266215)(156,37.30605244235137)(157,38.83100556226857)(158,34.777207018314066)(159,34.71950746884252)(160,35.01038719998463)(161,34.84484017915868)(162,35.08827768089082)(163,34.42928803522715)(164,34.755530184077614)(165,38.0122404870791)(166,26.95888224416178)(167,35.62260200947458)(168,34.191483186341436)(169,32.15748752936905)(170,35.078300659658794)(171,37.92850233923835)(172,33.61375077816532)(173,33.744568940947666)(174,27.949050384172825)(175,32.34950218626407)(176,31.177686816454553)(177,29.099397314049916)(178,31.388293377102265)(179,35.41279627322752)(180,30.185668888782978)(181,29.78185091274845)(182,31.57596951136443)(183,35.85079273124855)(184,31.50393909849164)(185,31.22862980796248)(186,31.258870636721984)(187,28.67130992850821)(188,30.610499908008034)(189,30.65632389667944)(190,33.119227494581594)(191,29.046708921771604)(192,29.782614555122194)(193,31.476749944497364)(194,30.024165526991233)(195,29.68624485488726)(196,30.181862654700772)(197,29.748169968254448)(198,29.95878589610469)(199,35.14728658915459)(200,28.129899011537695)(201,26.621881258344036)(202,28.00545108594114)(203,28.146858209887153)(204,27.73418915168782)(205,28.6430772246209)(206,27.685847568676504)(207,28.431880021438573)(208,28.282418761081843)(209,28.23674047813148)(210,26.94730158366927)(211,28.729795797946597)(212,27.94485815418658)(213,32.36851951624848)(214,33.817507935287274)(215,29.749250481919482)(216,29.638366545487674)(217,29.204553354605853)(218,32.219938734584375)(219,31.526905202474357)(220,30.724708493547723)(221,29.600010697135097)(222,28.938659145885303)(223,29.26316782059521)(224,30.536893300605342)(225,32.39049418984714)(226,34.0301918424442)(227,34.32932413599946)(228,34.171338928315734)(229,32.06488995473987)(230,31.921765792662114)(231,31.30704111119437)(232,30.383046259866152)(233,32.60653367923587)(234,31.39875456941633)(235,30.60232478379106)(236,30.57125804696905)(237,31.179979419655957)(238,31.102294979461565)(239,31.02690017153662)(240,29.995326657044302)(241,30.284972341377774)(242,30.475904665961117)(243,30.91234343637231)(244,31.994298654251963)(245,32.38267240910384)(246,32.85650573159482)(247,32.9876324794883)(248,33.29235141022233)(249,32.713215409121894)(250,32.843374024262154)(251,31.302604803275784)(252,32.86183498630029)(253,32.32519780378728)(254,32.8575661455314)(255,32.43604829481761)(256,33.21596416903663)(257,31.548079953086745)(258,33.236747904016354)(259,32.59783497345722)(260,33.77391021739323)(261,32.43531120121581)(262,32.92343685734848)(263,33.47107265686159)(264,32.850875842154736)(265,33.35262276225033)(266,33.71467893213307)(267,32.18393473250009)(268,31.525996062889206)(269,31.462066739837777)(270,31.976011975114414)(271,31.52916900580298)(272,32.51335813181309)(273,32.28402669446142)(274,33.30136255347282)(275,32.686113445180254)(276,32.94761257861281)(277,31.29579041870137)(278,31.671803532098416)(279,34.10122705308502)(280,31.833680710046277)(281,32.374550670934056)(282,31.070462874494314)(283,31.49245531610302)(284,32.4961099665863)(285,32.90577368299837)(286,31.7749704873386)(287,32.98900498499561)(288,34.17871190232177)(289,33.62645820233554)(290,33.70579089479425)(291,34.34726392646307)(292,31.592949641080793)(293,32.43879858532085)(294,32.86243870239172)(295,34.60635935488594)(296,33.30327542622085)(297,33.72137342114791)(298,28.860809721434354)(299,28.928535913174187)(300,29.764279077543797)(301,32.17801575035464)(302,32.24080176112186)(303,32.6546944550332)(304,30.439329716464425)(305,29.011517095523995)(306,28.566699933082376)(307,28.90003718146117)(308,30.851201770611134)(309,29.876372495419762)(310,30.294145364246894)(311,30.19949208683695)(312,31.780399477348475)(313,32.04386800815891)(314,33.40713742780071)(315,31.64094031163575)(316,29.76693038188828)(317,27.648106449635627)(318,33.08854282558698)(319,32.002410145881825)(320,30.93987777751552)(321,30.632665498040218)(322,30.078987950031575)(323,30.00805490487584)(324,28.428870257811642)(325,34.01665954820669)(326,31.90355961466398)(327,23.66135137600113)(328,29.29934314846583)(329,31.20258345704927)(330,32.051777337476004)(331,29.422431998173728)(332,27.074264370453704)(333,34.850572488726606)(334,31.629065956588498)(335,31.53604298256936)(336,36.68434521755207)(337,32.36161066691921)(338,30.146973311510287)(339,31.56481208638195)(340,31.65943907916826)(341,30.245478498761756)(342,30.259395002199376)(343,29.11692723066355)(344,31.725246773034392)(345,28.097073351803935)(346,27.863653972989734)(347,29.53060587021648)(348,28.54545303065887)(349,31.41810991883677)(350,28.519067429972093)(351,29.173908339738144)(352,31.29813724897278)(353,32.4211805221135)(354,30.775432265022637)(355,31.839487787360266)(356,30.055310148102382)(357,31.908161338051197)(358,30.097677135094997)(359,31.832202869636355)(360,29.841041853730687)(361,30.559208606113387)(362,29.732709929799466)(363,30.344311065497607)(364,33.45352677543415)(365,30.56256460202229)(366,32.05329696753703)(367,30.689405786602425)(368,29.621224047015506)(369,29.34000316066193)(370,30.195869141945238)(371,28.776736208285943)(372,27.529511656405496)(373,27.216253342693864)(374,25.93552512584396)(375,31.85710608970847)(376,30.40162762966398)(377,31.312383226263936)(378,22.569046784698948)(379,23.524282029594062)(380,30.425248104960247)(381,22.956224378451306)(382,33.883875101217534)(383,32.27267440049514)(384,29.209102146705884)(385,30.91459524200088)(386,22.815464732007285)(387,23.412462691092557)(388,27.11801339368249)(389,19.633341838366082)(390,20.51588118919333)(391,29.58650383724474)(392,25.952032447947484)(393,30.499098583037167)(394,26.339010500901527)(395,32.87271791883963)(396,25.543408656946948)(397,27.980465291949365)(398,27.51539134361204)(399,27.494532089969226)(400,32.69310598065523)(401,28.256018543020705)(402,28.16647440647951)(403,29.155310093451813)(404,28.36059019880459)(405,27.709929763143393)(406,27.233508908504827)(407,27.598943816629156)(408,31.255887627478504)(409,26.86874148609336)(410,27.688858909501686)(411,27.822838955327548)(412,29.031558151447655)(413,29.53860823564048)(414,29.732153481734723)(415,29.160625378731517)(416,30.726573186168377)(417,32.41713246267153)(418,30.092011210180125)(419,29.665926442792745)(420,30.109709437510197)(421,30.838352698462245)(422,28.64986184098351)(423,25.85628930857473)(424,31.717403944224614)(425,36.48134469322633)(426,32.23773490506026)(427,31.291788745961966)(428,31.567267980771668)(429,30.51552371477598)(430,31.158050693134406)(431,30.15983732796977)(432,29.771520538165817)(433,30.821565659612244)(434,31.856969819537895)(435,30.50046423064994)(436,30.27691717193806)(437,31.32279983601437)(438,31.04753432589877)(439,30.87562431103947)(440,29.108031678123478)(441,31.435954023541953)(442,30.442066039627907)(443,29.599625207120823)(444,28.650817579415733)(445,29.289850282541718)(446,29.38279778203546)(447,33.153773432220056)(448,32.62741372562545)(449,30.649283370453492)(450,30.798779177682558)(451,32.50023681762542)(452,31.836011720691314)(453,30.484790282221518)(454,33.08667985321218)(455,30.65946273559983)(456,32.95064701178661)(457,29.959671100255044)(458,32.477722050401574)(459,31.55356280076039)(460,31.148253316346743)(461,31.560526808282994)(462,30.7413326240644)(463,32.36713902909891)(464,30.43472826601993)(465,32.84177528484175)(466,33.583904093194114)(467,33.08074506956381)(468,31.321517307567493)(469,30.25637631744192)(470,33.59622839145533)(471,30.779223408897252)(472,30.23252520828045)(473,30.573076688210783)(474,30.508823647861064)(475,33.67900676176632)(476,31.588751989883136)(477,35.54488456930234)(478,33.307379779731306)(479,33.94676589171031)(480,31.60014433077903)(481,31.057804065647442)(482,30.11399463127453)(483,29.513103636174964)(484,31.666543018596194)(485,26.28938848960188)(486,28.665029559688342)(487,29.709638327342084)(488,29.588636657378775)(489,29.194470298574835)(490,29.161652122542545)(491,29.315139271089127)(492,31.1914386907379)(493,31.803222533062574)(494,28.53982962831062)(495,28.955014767643863)(496,31.623531722463785)(497,33.75557135844958)(498,29.765165403132695)(499,31.22697041258441)(500,28.535436069137976)(501,26.051653071204814)(502,28.60238033836448)(503,32.245140736825405)(504,38.715632781799435)(505,30.658090930314806)(506,29.115695486621316)(507,37.01925794442354)(508,32.561215385909584)(509,27.651329973828695)(510,24.659109897580787)(511,28.52466640702869)(512,29.570819117285865)(513,33.25203995593075)(514,30.48023140553708)(515,30.03950484572826)(516,30.90254654000107)(517,27.556513743607617)(518,31.860493025738993)(519,27.339904363883388)(520,28.6020555780468)(521,32.76181340415114)(522,29.65741487713692)(523,30.26558784007498)(524,29.249413075483037)(525,30.34735601249918)(526,27.711756275642525)(527,29.238825468633763)(528,29.706445392156844)(529,28.707144463500633)(530,28.376735739228113)(531,28.296788218483012)(532,30.010449369389118)(533,29.393486146802996)(534,32.29296767219267)(535,28.228005954048538)(536,29.78361619856359)(537,29.909509715266406)(538,27.663200594947657)(539,27.934396762275266)(540,28.80486548985591)(541,29.334004833335737)(542,29.388272828138547)(543,27.938274100498305)(544,31.454250083631365)(545,29.431339981190973)(546,28.454581556865616)(547,31.285550165331006)(548,33.92302657546514)(549,25.542590535986392)(550,28.420521997936213)(551,29.486031423567702)(552,29.201022482528472)(553,28.30209395387946)(554,28.388006141434698)(555,27.482441449376356)(556,33.764025737002555)(557,24.194803725284267)(558,29.927498420057756)(559,30.38150907951358)(560,29.885041464598313)(561,25.80903382351432)(562,30.45955187640662)(563,30.349958880345973)(564,27.47622665017212)(565,29.64329741875276)(566,29.143280820516953)(567,30.212049015248606)(568,32.880641691817765)(569,30.741282441711405)(570,23.23251422986174)(571,29.64164536613819)(572,31.58875952746511)(573,32.61580671605378)(574,32.32282564525808)(575,32.785433060125044)(576,28.83925476233329)(577,28.045656735780756)(578,31.276117726990282)(579,30.60281123860573)(580,28.613872718491738)(581,31.35996339667135)(582,32.380985896089)(583,32.75559846641012)(584,32.49502144038769)(585,32.62205714689074)(586,29.498433798439795)(587,31.310693135005675)(588,30.207126924918104)(589,32.96896595925641)(590,30.63517897571769)(591,31.72632126724885)(592,31.77632854347767)(593,32.07400070372016)(594,30.179711006476747)(595,30.152725192741084)(596,30.305048489954547)(597,27.833612226002863)(598,28.50295401317931)(599,29.237346110230163)(600,25.338421734843543)(601,29.827346019725848)(602,28.685196161555684)(603,30.820457277179386)(604,27.67696041016362)(605,29.71055522602114)(606,27.577446052006707)(607,32.579338392426216)(608,29.254212191518878)(609,28.758917140367423)(610,31.07818549795075)(611,28.58780346084688)(612,31.175191403094626)(613,28.76360729152173)(614,31.68924648841792)(615,28.472660541146936)(616,29.613360835461563)(617,28.571200085804307)(618,30.507573520915717)(619,29.190277153302198)(620,28.301092394414724)(621,29.802013355603687)(622,31.045888083343026)(623,27.565931964518846)(624,28.914274551902626)(625,27.863158946413048)(626,27.68346263229564)(627,30.906599220320434)(628,28.181288944405146)(629,28.808212686248282)(630,29.838179622516776)(631,27.87077137027918)(632,27.34933024518596)(633,31.08219652587415)(634,28.0933581839587)(635,28.366597620731387)(636,28.307905518963683)(637,27.450977575102456)(638,29.498886969786653)(639,27.869665158857902)(640,32.74939932605861)(641,28.405537748079603)(642,29.43575674410942)(643,27.392756479563445)(644,30.07712678781366)(645,29.535967141027857)(646,27.65402618891492)(647,28.662318465358464)(648,30.856558034637445)(649,30.18292476103776)(650,30.50917569420267)(651,31.122626454064708)(652,30.528638509453703)(653,30.922713972587978)(654,30.960810122569505)(655,27.168144990740622)(656,29.95937977312227)(657,30.715269237672256)(658,30.107456835520406)(659,30.728753133762645)(660,29.58382727135874)(661,29.01845553142696)(662,28.51065029725075)(663,30.581324387928444)(664,27.5659299948952)(665,28.234455825629972)(666,32.74541368022398)(667,27.87994089708905)(668,29.255782762939276)(669,28.40144649294826)(670,31.223564880003853)(671,30.757066251059243)(672,33.294067876372104)(673,34.471695368936864)(674,31.042054193964628)(675,34.90886653471979)(676,32.706757733118124)(677,33.75376751352945)(678,38.211837938435366)(679,31.3064853055578)(680,36.43399750797823)(681,33.57104561012504)(682,31.129315669375405)(683,30.37042437531325)(684,33.79077102811124)(685,33.2160188777861)(686,31.581401806058263)(687,34.45506435139538)(688,37.37695228771978)(689,32.2475061402151)(690,35.189391872967086)(691,31.088005470843306)(692,36.51317046060269)(693,35.91372314704302)(694,32.29472081803708)(695,34.21060550621034)(696,30.061019827639395)
    };

\end{axis}
\end{tikzpicture}

%% file: method.tex
\section{Method}\label{sec:method}

In this section, we first describe the self-supervised training paradigm used by state-of-the-art monocular depth estimation methods~\cite{monodepth17, monodepthv2}.
Then, we discuss how current methods deal with the issue of scale ambiguity by accessing ground truth 3D information at test time.
Finally, we detail how basic prior information on a vehicle-mounted can be used to calibrate self-supervised depth map predictions to produce depth map with a correct physical scaling without requiring the acquisition of 3D or other information by means of additional sensors.

\subsection{Self-supervised Depth Estimation}\label{s:self-sup}

\begin{figure}[b]
    \centering
    \includegraphics[width=\textwidth]{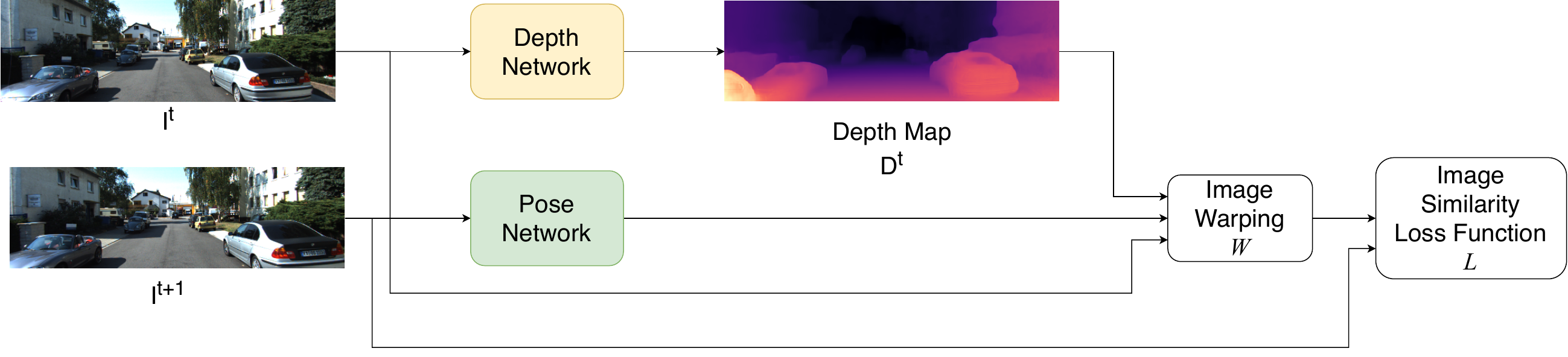}
    \caption{Self-supervised monocular depth estimation pipeline~\cite{monodepth17, monodepthv2}
    \vspace{-15pt}
    }\label{fig:architecture}
    \vspace{-10pt}
\end{figure}

Given a pair of subsequent video frames $I^t$ and $I^{t+1}$ captured by a moving camera, under mild conditions such as Lambertian reflection, the image $I^t$ is (approximately) a warp (deformed version) of image $I^{t+1}$~\cite{Hartley2004}.
Moreover, the warp depends only on the \emph{geometry and motion} of the scene, captured by the depth map $D_t$ and the viewpoint change $(R^t,T^t)$.
In other words, we can write $I^t \approx \mathcal{W}(I^{t+1}, D_t, R_t, T_t, K)$, where $\mathcal{W}$ is a warp~\cite{JaderbergSZK15} which depends only on the depth $D_t$, the viewpoint change $(R_t, T_t)$, and the camera intrinsics $K$ (which we assume known and constant).

The equation above provides a constraint that can be used to self-supervise a monocular depth estimation network $\Phi$ from knowledge of the video frames $I^t$ and $I^{t+1}$ alone.
In more detail, we task two networks $\Phi$ and $\Psi$ to predict respectively the depth $D^t = \Phi(I^t)$ from the first image and the motion $(R_t, T_t) = \Psi(I^t,I^{t+1})$ from the pair of images so as to correctly warp $I^{t+1}$ into $I^t$, thus establishing the expected visual consistency (see \cref{fig:architecture}).
This is done by minimizing the appearance loss between the original $I^t$ and the synthesized image $\hat{I}^t = \mathcal{W}(I^{t+1}, D_t, R_t, T_t, K)$:
\begin{gather}
\mathcal{L}(I^t, I^{t+1}) = \alpha E_{\mbox{\scriptsize p}}(I^t, \hat{I}^{t}) + E_{\mbox{\scriptsize dis}}(I^t, \hat{I}^{t}) \label{eq:lossFunction}\\
\hat{I}^t = \mathcal{W}(I^{t+1}; D_t, R_t, T_t, K),
~~~~
D_t = \Phi(I^t),
~~~~
(R_t, T_t) = \Psi(I^t, I^{t+1})
\end{gather}
The photometric loss term $E_{\mbox{\scriptsize p}}$ in \cref{eq:lossFunction} is the SSIM loss~\cite{wang2004image,lossfunctions}, whilst the $E_{\mbox{\scriptsize dis}}$ term enforces smoothness~\cite{monodepth17}.
The whole network is trained end-to-end using standard back-propagation.

An analysis of the warp operator $\mathcal{W}$~\cite{JaderbergSZK15} shows that the operator is invariant to multiplying the depth and the translation parameters by a constant $\alpha$:
\begin{align}
\mathcal{W}(I; D, R, T, K) = \mathcal{W}(I; \alpha D, R, \alpha T, K)\quad \alpha\in\mathbb{R}.
\end{align}
This shows that the network can only learn depth and translation up to an undetermined scaling factor $\alpha$;
in particular, there is no reason for the learned scale to corresponds to the true physical scale of the scene.
As a matter of fact, the model is not even forced to learn a scaling factor consistently across different pairs of frames $(I^t,I^{t+1})$, which we show in empirically is in fact not the case.
In particular, \cref{fig:monodepthscalingfactor}~shows that the variation in scaling factor for different frames can be up to a factor of two.

\subsection{Ground Truth Data used to Scale Depth at Test Time}\label{sec:gtscaling}

Since the scale of the predicted depth $\Phi(I^t)$ is arbitrary, its use in downstream tasks that require a physical understanding of the scene (e.g.~in robotics) impossible.
Equally, all benchmarks for depth estimation~\cite{Geiger2012CVPR} also require measurements in real units (meters), and therefore the depth map predictions $\Phi(I^t)$ \emph{cannot} be assessed directly against these benchmarks.
Instead, the common approach is to just marginalize out the scale at test time, finding the factor that best matches the predicted and ground-truth depth \emph{for each test image independently}~\cite{monodepth17,casser2018depth, luo2018every,monodepthv2}.
Since this ground-truth information is obtained via a sensors such as a LiDAR, this is equivalent to calibrating the method against an additional sensor, which is not a realistic setup.

More formally, given an image $I$, the network outputs prediction a $\Phi(I)$ which is transformed to the final depth estimate as $d_I = \alpha_I\Phi(I)$ where:
\begin{equation}\label{e:lidar-calibration}
\alpha_I =
\frac
{\operatorname{median} d_I^{\mbox{gt}}}{\operatorname{median} \Phi(I)}
\end{equation}
where $d_I^{\mbox{gt}}$ is the ground-truth depth map, usually created by projecting sparse LiDAR points onto the image plane, projected with the same viewpoint as the input image $I$.\footnote{Both images are masked such that the scaling factor is only calculated on points where the LiDAR has read data}

\subsection{Road Model Estimation}\label{sec:road_model_est}

In order to remove the need for LiDAR ground truth at test time, we exploit prior knowledge of the environment and of the camera setup, especially the camera height. Because cars drive on roads and we know that the camera is at certain height above the road, we can exploit this constraint to calibrate the depth map to real-world values.

In order to do so, we first need to automatically estimate a road model in every test image.
In order to account for the fact that many roads are not perfectly flat, more typically they slope up/down or are higher on one side than the other, or that the car tilts during acceleration and deceleration, we estimate the pitch and roll of the road by fitting a plane to the raw depth map. We only use values for pixels that are classified as road by a pre-trained semantic segmentation model~\cite{zhou2018semantic}, and whose $|X|$ and $Z$ co-ordinate\footnote{The world co-ordinates $X, Z$ to filter out pixels above the threshold are obtained using camera intrinsics and assuming the road is perfectly flat, i.e. $Y = -h$} is below a certain threshold (see \Cref{sec:ablations}). 

We then fit these points using Least Median of Squares regression~\cite{rousseeuw1984least} to get the road plane estimate in the 3D world $a_1X + a_2Y + a_3Z + c = 0$.
We know that the 3D point on the road right below the camera has the co-ordinate $[0, -h, 0]$, where $h$ is the camera height, and therefore we can infer the following relation for the scaling factor
\begin{align}\label{e:our-calibration}
\alpha_I &= \frac{c}{h}
\end{align}
Compared to~\cref{e:lidar-calibration}, the scaling factor~\cref{e:our-calibration} now only relies on the visual information, and therefore can be obtained without any LiDAR input.


\begin{figure}
\centering\noindent
\footnotesize
\begin{tabular}{p{.3\textwidth}p{.61\textwidth}}
\mbox{}
\newline \includegraphics[width=0.3\textwidth]{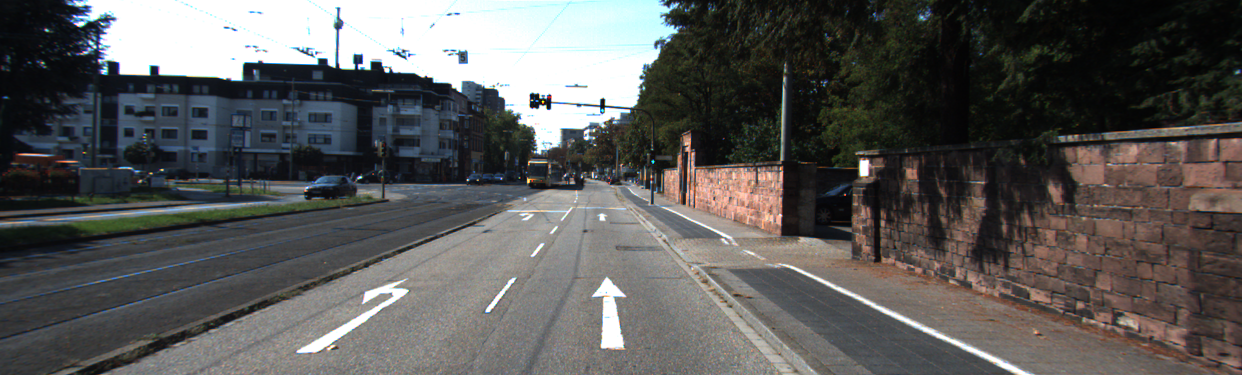}%
\newline Input image $I$
\newline \includegraphics[width=0.3\textwidth]{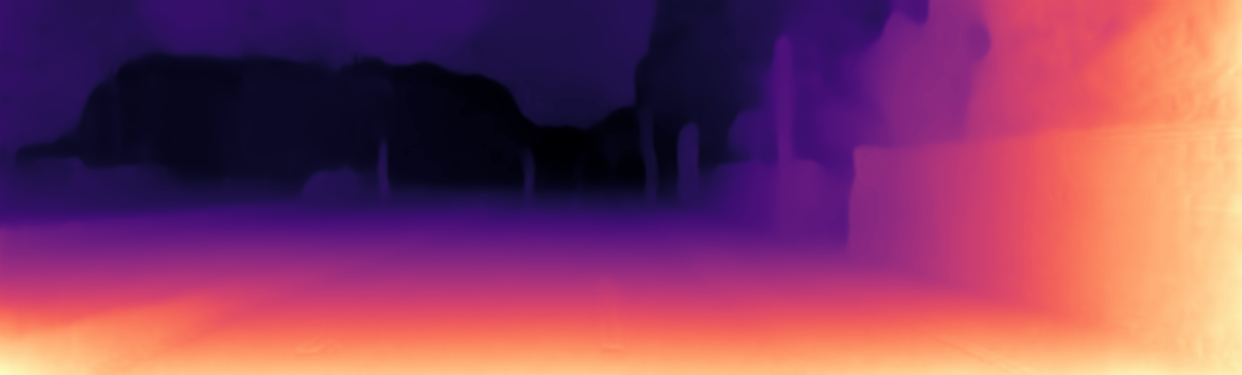}
\newline Raw output $\Phi(I)$
\newline \includegraphics[width=0.3\textwidth]{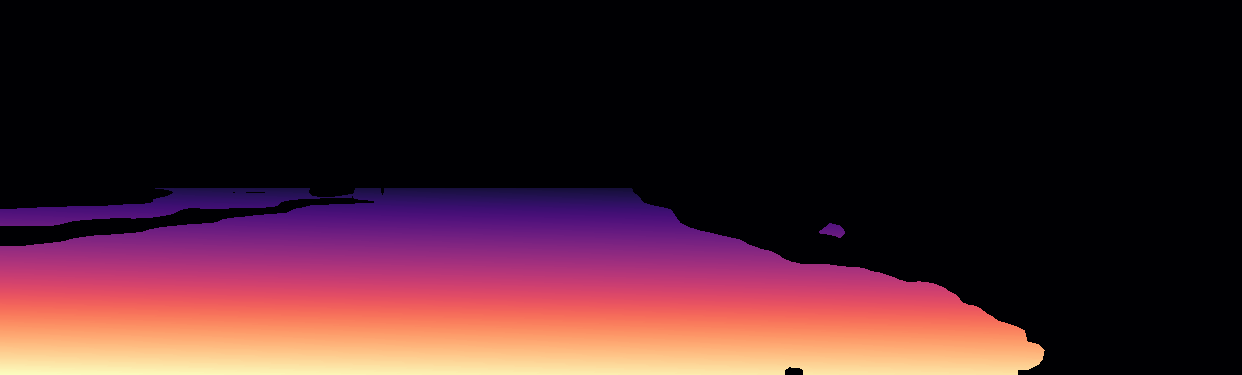}
\newline Restricted to road pixels
&
\mbox{}
\newline \includegraphics[width=0.61\textwidth]{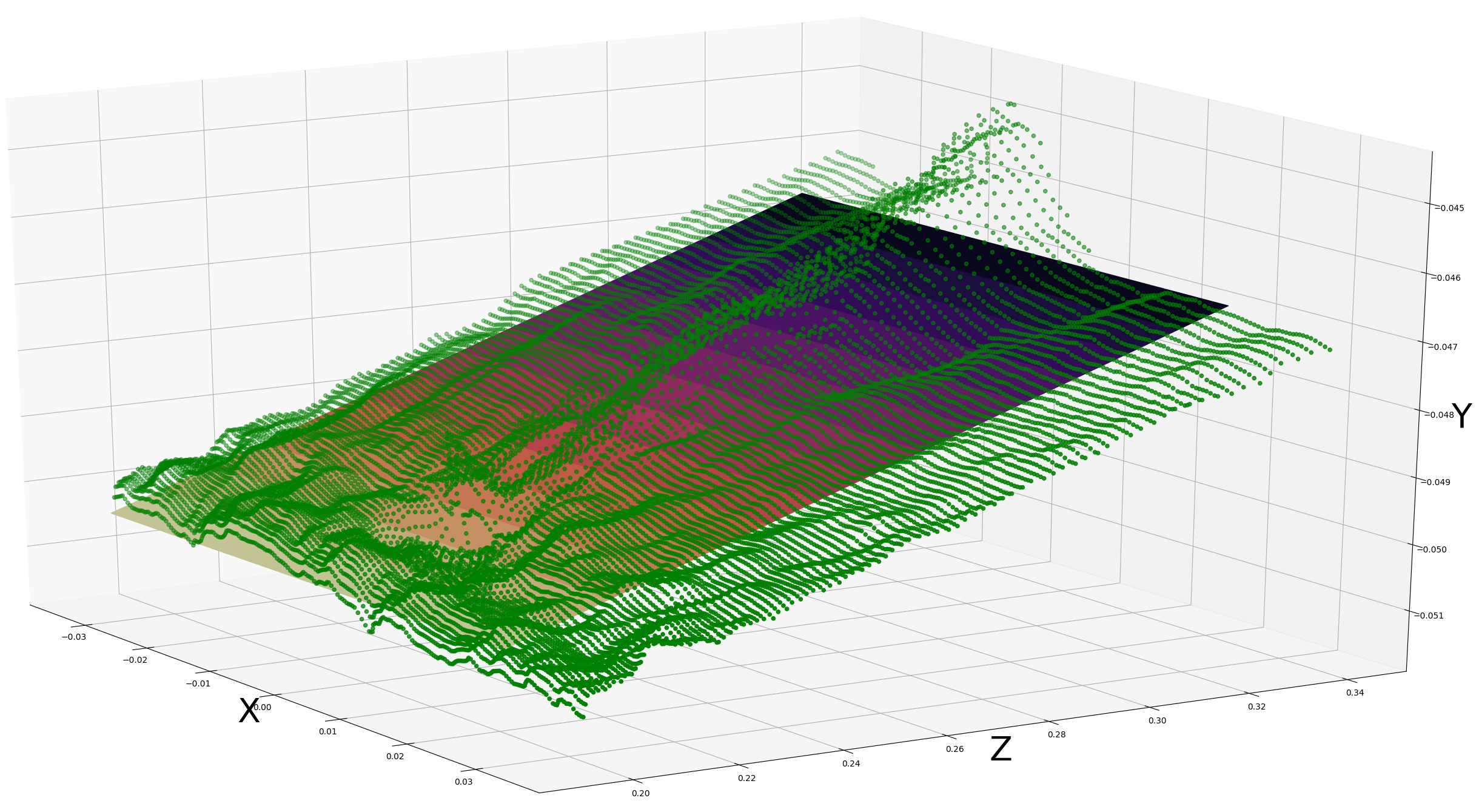}
\newline Parametric fit
\end{tabular}

\caption{Road model estimation. First we take the uncalibrated depth of the input image, combined with scene segmentation to extract only depth values belonging to the road. After further refinement, we project the points into 3D and fit a plane to them \vspace{-10pt}}\label{fig:pseudo-road}
\vspace{-10pt}
\end{figure}

\input{figs/experiments/qualitative}

%% file: figs/experiments/qualitative.tex
\newcommand{\turnheightnew}{0.40\columnwidth}
\begin{figure}[ht!]
    \centering
    \begin{tabular}{@{\hskip 1mm}c@{\hskip 1mm}c@{\hskip 1mm}c@{}}

        {\rotatebox{90}{\hspace{4mm}\footnotesize Input}} 
            & \includegraphics[width=\turnheightnew]{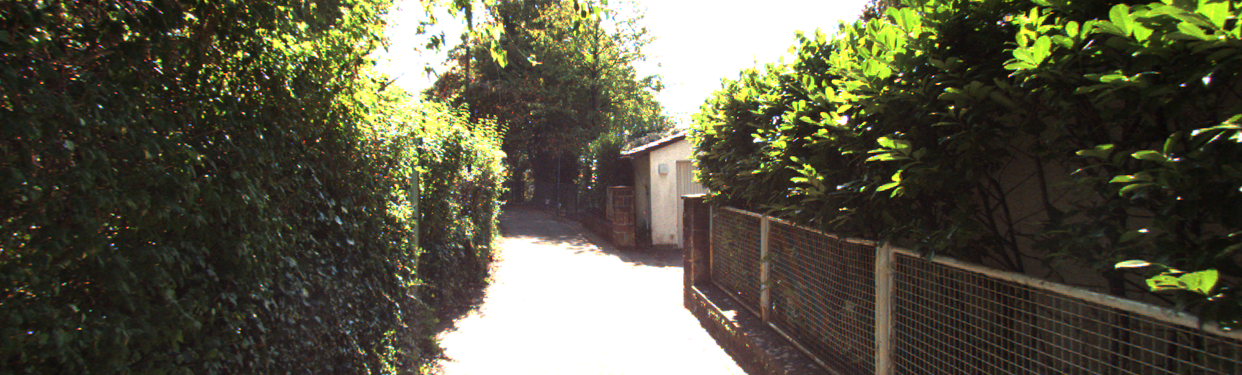} 
            & \includegraphics[width=\turnheightnew]{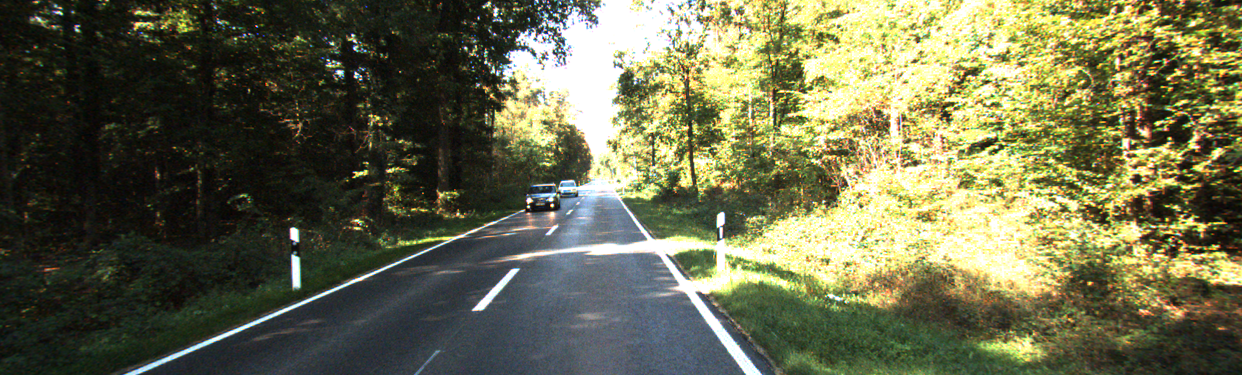} 
        \\  
        
        {\rotatebox{90}{\hspace{0mm}\footnotesize Raw Output~\cite{monodepthv2}}} 
            & \includegraphics[width=\turnheightnew]{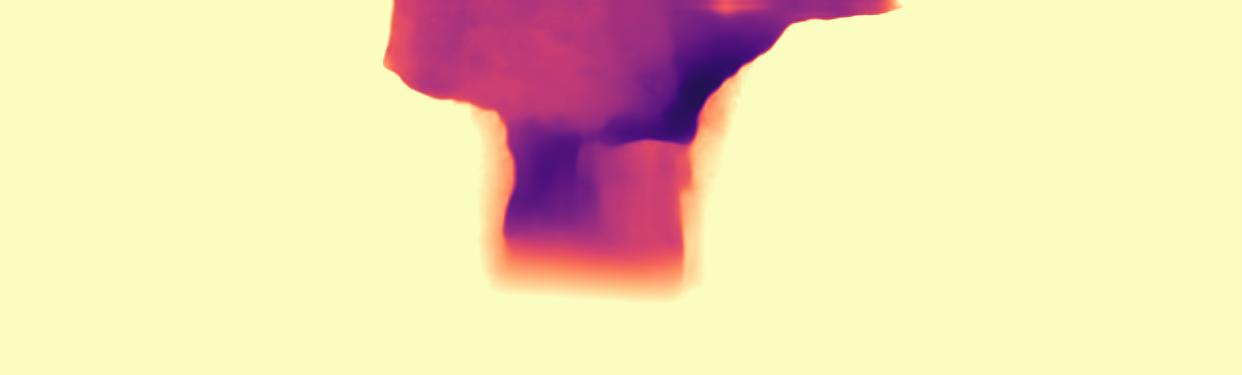} 
            & \includegraphics[width=\turnheightnew]{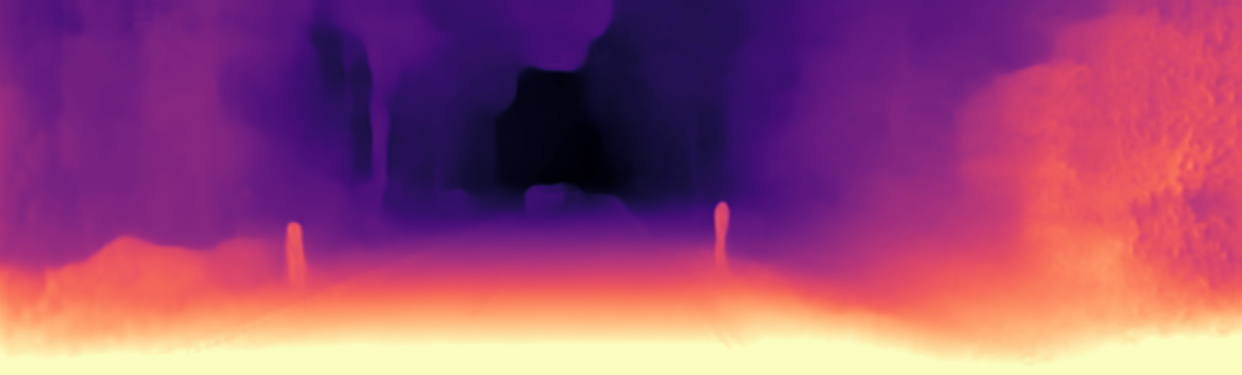} 
        \\
        
        {\rotatebox{90}{\hspace{1mm}\footnotesize GT Scale~\cite{monodepthv2}}} 
            & \includegraphics[width=\turnheightnew]{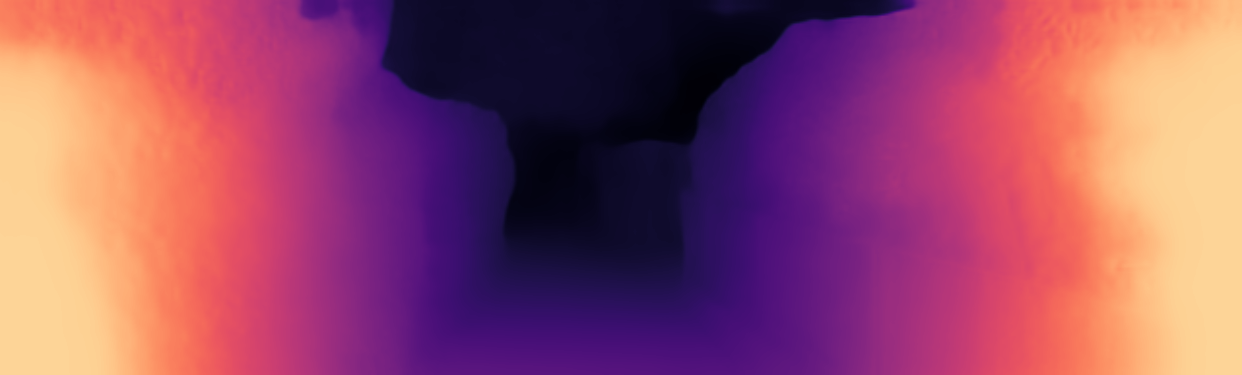} 
            & \includegraphics[width=\turnheightnew]{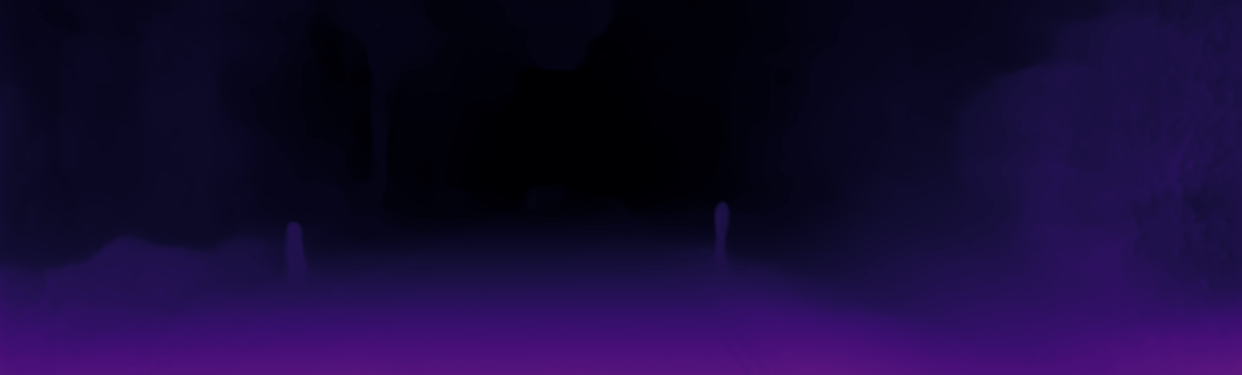} 
        \\

        {\rotatebox{90}{\hspace{2mm}\footnotesize Single Scale}}
            & \includegraphics[width=\turnheightnew]{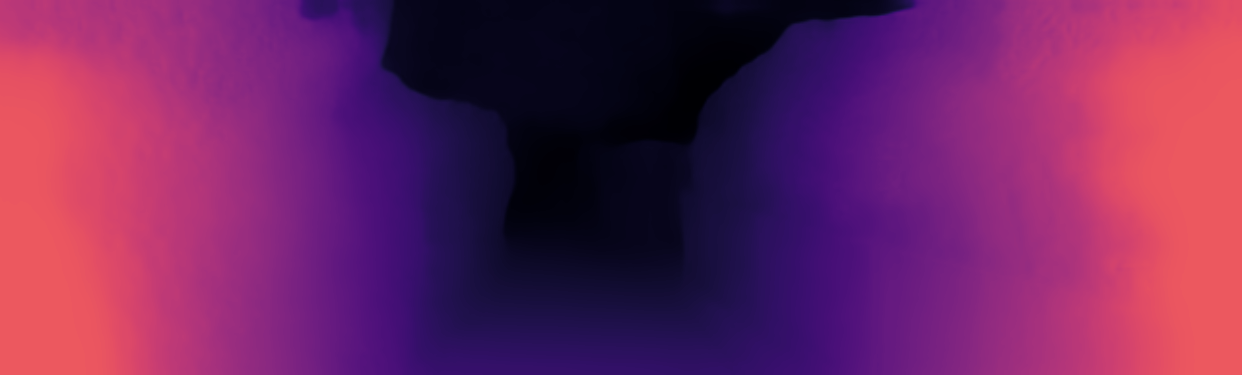} 
            & \includegraphics[width=\turnheightnew]{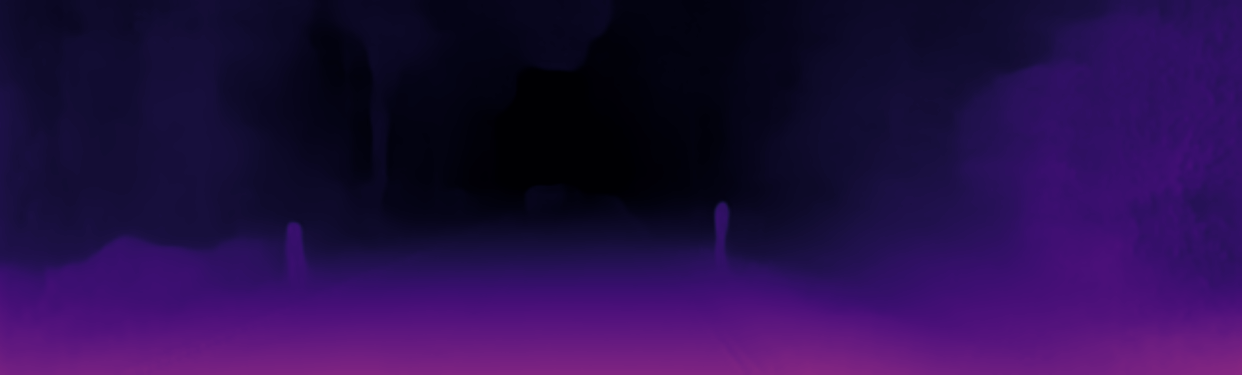} 
        \\
        
         {\rotatebox{90}{\hspace{5mm}\footnotesize ours}}
            & \includegraphics[width=\turnheightnew]{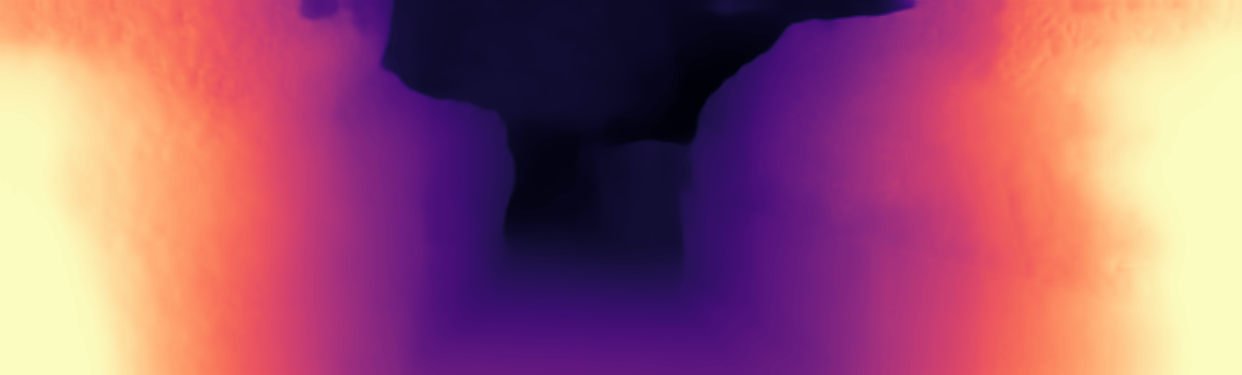} 
            & \includegraphics[width=\turnheightnew]{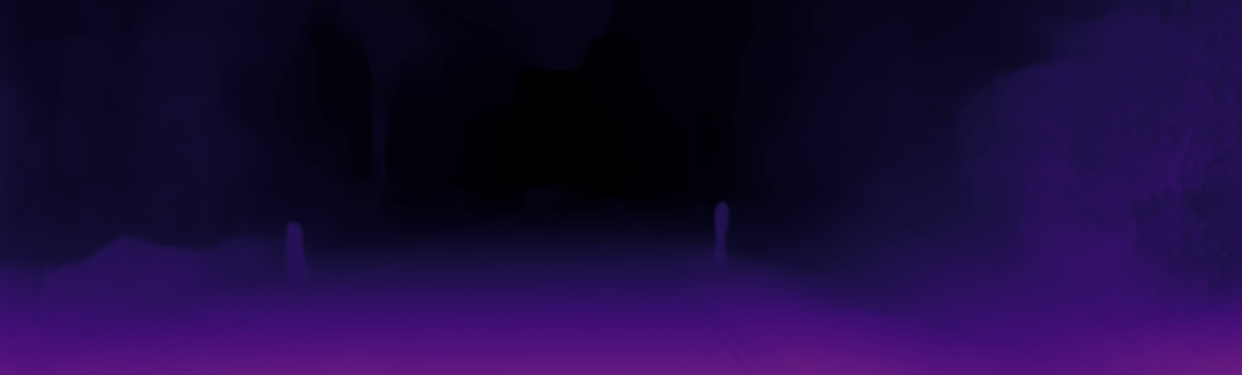} 
        \\

        {\rotatebox{90}{\hspace{6mm}\footnotesize GT}} 
            & \includegraphics[width=\turnheightnew]{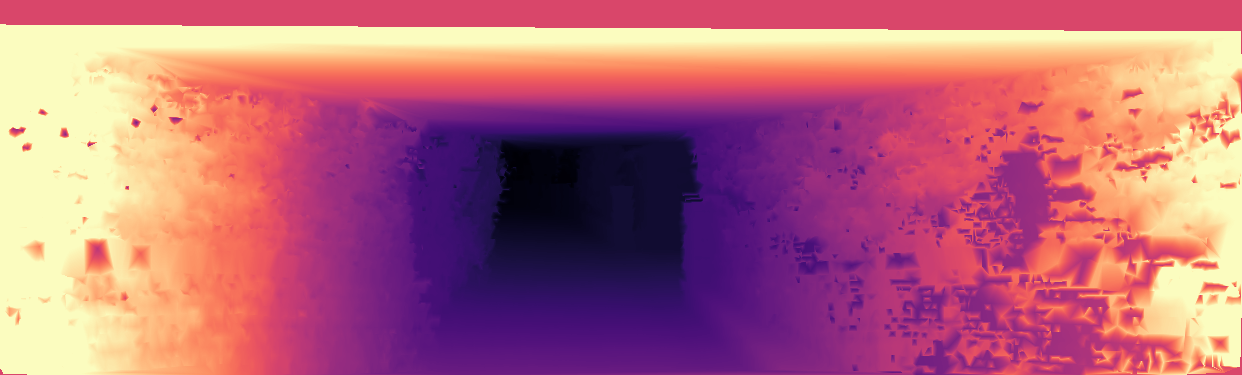} 
            & \includegraphics[width=\turnheightnew]{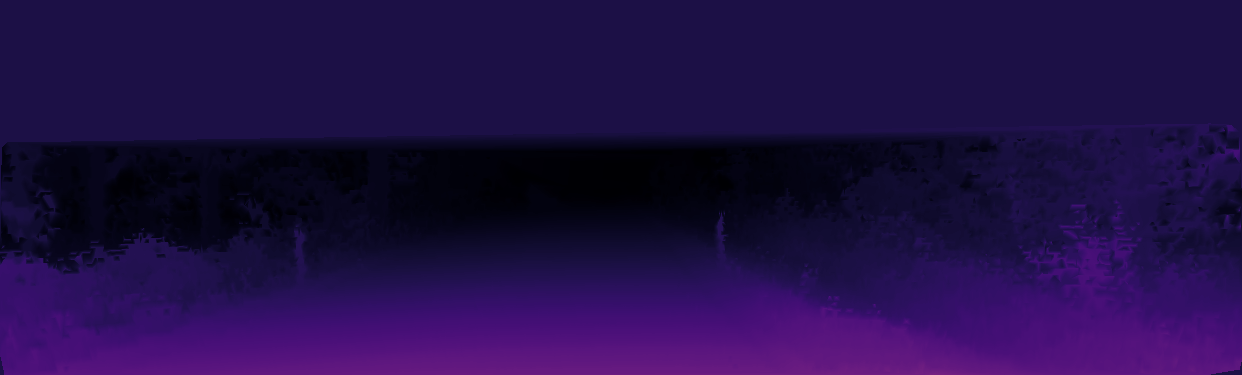} 
        \\
        {
        \rotatebox{90}{\hspace{0mm} \footnotesize {Distance [m]}}}
        & \multicolumn{2}{c}{\includegraphics[width=\turnheightnew]{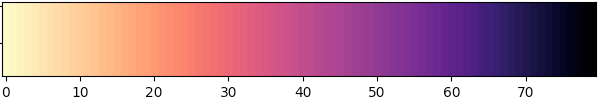}}

        \end{tabular}
        
    \caption{Qualitative depth estimation examples from the KITTI dataset (inverse depth shown). Monodepth2~\cite{monodepthv2} output values (Raw Output) are scaled by comparing the output to the ground truth for every test image (GT Scale). Using a single scaling factor from the training set (Single Scale) is significantly worse. Using road model (ours) to estimate the scaling factor achieves significantly better results. All images use the same color coding.\vspace{-10pt}}
    \label{fig:qualitative}
    \vspace{-10pt}
\end{figure}

%% file: experiments.tex
\section{Experiments}\label{sec:experiments}

In this section, we compare our calibration method (Ours) to:
(1) the un-calibrated outputs of the network $D=\Phi(I)$ (Raw);
(2) computing a per-frame calibration factor using \cref{e:lidar-calibration} with access to the 3D ground-truth (GT Scaling); and
(3) the same as (2), but by computing a single scaling factor from either all the training or testing frames (GT Single Scaling).
After a qualitative and quantitative comparison with these techniques and state-of-the-art monocular depth estimation networks (both supervised and unsupervised), we ablate our method, showing the importance of the various components, and study sensitivity to its parameters.


\paragraph{Implementation details.}

In all our experiments, we used the MonodepthV2~\cite{monodepthv2} pre-trained model. In line with prior work, we use the Eigen et al. \cite{EigenF14} data split of KITTI dataset~\cite{Geiger2012CVPR}.

\subsection{Qualitative comparison}

We first look at the optimal scaling factor determined via GT Scaling via access to the ground-truth at test time (\cref{sec:gtscaling}, \cite{monodepth17,monodepthv2}).
As is observable in \Cref{fig:monodepthscalingfactor}, the factor selected by GT Scaling varies wildly even in a single video sequence.
This is illustrated in \cref{fig:qualitative} for two different input images.
GT Scaling chooses factor 19.63 for the left image vs GT Single Scaling (median on the training set) of 30.462.
This means that, for this image, the network predicts a depth map where objects are 50\% farther away than for the median case.
The image to the right is the opposite, as GT Scaling determines the best factor to be 40.55, so objects are predicted to be 33\% than the median case.
Given these differences, it is clear that there is no single scaling factor that results in a good fit for all test frames; hence, below we find it unsurprising that using a single scaling factor over the entire test set (GT Single Scaling) produces inaccurate results overall.

By comparison, our scaling technique predicts scaling factors of 19.63 and 36.4 for the two images respectively, which are close to the output of GT Scaling.
Hence, our system produces results significantly closer to the per-frame GT Scaling factors than GT Single Scaling while having \emph{no} access to ground-truth (LiDAR) 3D information at training or test time.
This useful for autonomous vehicles that wish to adapt to scenes where it frequently drives rather than examples in a training set as is done in \cite{mccraith2020monocular}.

\subsection{Quantitative comparison}
\input{tables/misc} 
\input{tables/quantitative}

First, in \cref{tab:naive} we contrast our method (d) to GT Single Scaling, fixing the scaling factor using respectively the training and testing subset of the data (a) and (b).
We note that our approach is substantially better than both (0.113 vs $\geq$ 0.125 AbsRel).
This is because, while GT Single Scaling has access to 3D ground-truth, it uses a fixed scaling factor for all frames, and, as shown above, no single scaling factor can work well.
Remarkably, our method is comparable to GT Scaling \emph{as well} (the latter corresponds to the penultimate row of~\cref{tab:kitti_eigen}), matching it, in particular, in the Abs Rel metric, despite the fact that GT Scaling chooses the best possible scaling factor for each frame individually against the ground-truth.
From the same table, we see that this is obtained against a model, Monodepth~V2, which is state-of-the-art, resulting for the first time in excellent \emph{calibrated} self-supervised monocular depth estimation from \emph{vision alone}.

\subsection{Ablation and tuning}\label{sec:ablations}

\input{tables/pseudo_road_len}
\input{tables/pseudo_road_width}

\paragraph{Road model.}

Recall that our method is based on estimating the full 3 DoF of the ground plane.
First, we test whether this is necessary.
In order to do so, we assume instead that the plane is exactly horizontal and at the fixed canonical height below the camera.
Then, we use the fixed plane to generate a pseudo-LiDAR map for the road pixels and use GT Scaling against those pseudo-ground-truth values (instead of the actual GT value) in order to determine the scaling factor for each frame.
A similar fixed pseudo-LiDAR plane was also used, for example, in Segment2Regress\cite{Segment2Regress} in order to perform 3D object detection.
The result of this is shown in \cref{tab:naive} row (c):
the fact that this simple fixed-plane model ignores the tendency of real roads to have inclines and declines as well as the cameras ability to have non-negligible pitch and roll during regular car motion which greatly effects it's depth prediction in the far range.

\paragraph{Tuning.}

Next, we assess the sensitivity of our methods to various parameters and determine their optimal values.
In \cref{sec:road_model_est} we first take to varying the maximum distance of points on the road we use for the plane fitting.
Similar to \cite{groundnet} we find that using points predicted to be under 30 meters from our camera works best for fitting our ground plane as seen in \cref{tab:road_width}.
In a similar fashion we explore the maximum left-right distance from which we consider points from our fit.
Typical roads are between 2.75 and 3.75 meters wide so it is within reason that only points within 3m left of right of the car work best with a gradual drop off in performance above this and an insufficient number of points below. Note that the method is not overly sensitive to a specific parameter setting, likely due to the use of robust estimator.




%% file: tables/misc.tex
\begin{table}[!t]
  \centering
  \caption{Comparison of different depth map scaling methods on the KITTI testing subset.}
  \resizebox{\textwidth}{!}{
  \begin{tabular}{|l|c|c|c|c|c|c|c|c|}
  \hline
  Method  & \cellcolor{col1}Abs Rel & \cellcolor{col1}Sq Rel & \cellcolor{col1}RMSE  & \cellcolor{col1}RMSE log & \cellcolor{col2}$\delta < 1.25 $ & \cellcolor{col2}$\delta < 1.25^{2}$ & \cellcolor{col2}$\delta < 1.25^{3}$\\
 
  \hline
(a) GT Single Scaling (training set) & 0.125 & 0.942 & 5.045 & 0.208 & 0.84 & 0.953 & 0.979 \\
(b) GT Single Scaling (testing set) & 0.126 & 0.952 & 4.999 & 0.204 & 0.848 & 0.954 & 0.98 \\
(c) Fixed road plane & 0.132 & 1.073 & 5.035 & 0.203 & \textbf{0.86} & \textbf{0.954} & \textbf{0.977} \\
\textbf{(d) \emph{Ours}} &  \textbf{0.113} &\textbf{ 0.916} & \textbf{4.974} & \textbf{0.199} & 0.857 & 0.945 & 0.968 \\
\arrayrulecolor{black}\hline
\end{tabular}}
\label{tab:naive}  
 \vspace{-10pt}
\end{table}

%% file: tables/quantitative.tex
\begin{table}[ht!]
\small
  \setlength{\tabcolsep}{2pt}

  \centering

  \begin{tabular}{|l|c|c||c|c|c|c|c|c|c|}
  \hline
  Method & Train & GT@Test & \cellcolor{col1}\footnotesize Abs Rel & \cellcolor{col1}\footnotesize Sq Rel & \cellcolor{col1}\footnotesize RMSE  & \cellcolor{col1}\footnotesize RMSE log & \cellcolor{col2}\footnotesize  $\delta<1.25 $ & \footnotesize \cellcolor{col2}$\delta<1.25^{2}$ & \cellcolor{col2}\footnotesize $\delta<1.25^{3}$\\
  \hline

Eigen~\cite{EigenF14} & D &\xmark & 0.203 & 1.548 & 6.307 & 0.282 & 0.702 & 0.890 & 0.890\\
Liu~\cite{LiuSLR15} & D &\xmark&  0.201 & 1.584 & 6.471 & 0.273 & 0.680 & 0.898 & 0.967\\
Klodt~\cite{klodt2018supervising} & D*M &\xmark&  0.166 & 1.490 & 5.998 & --- &  0.778 & 0.919 & 0.966\\
AdaDepth~\cite{gandepth2018}  & D* &\xmark&  0.167 & 1.257 & 5.578 & 0.237 & 0.771 & 0.922 & 0.971\\
Kuznietsov~\cite{KuznietsovSL17} & DS &\xmark&  0.113 & 0.741 & 4.621 & 0.189 & 0.862 & 0.960 & 0.986\\
DVSO~\cite{yang2018deep} & D*S &\xmark&  0.097 & 0.734 & 4.442 & 0.187 & 0.888 & 0.958 & 0.980\\
SVSM FT~\cite{singlestereo2018} & DS &\xmark& \underline{0.094} & \underline{0.626} & 4.252 & 0.177 & 0.891 & 0.965 & 0.984\\
Guo~\cite{guo2018learning} & DS & \xmark & 0.096 & 0.641  & \underline{4.095}  & \underline{0.168}  & \underline{0.892}  & \underline{0.967}  & \underline{0.986} \\
DORN~\cite{Fu_2018_CVPR} & D & \xmark & \textbf{0.072}&  \textbf{0.307} & \textbf{2.727} & \textbf{0.120} & \textbf{0.932} & \textbf{0.984} & \textbf{0.994}\\

\arrayrulecolor{black}\hline

Zhou~\cite{zhou2017}\textdagger & M & \cmark & 0.183 &   1.595 & 6.709 & 0.270 & 0.734 & 0.902 & 0.959\\
Yang~\cite{yang2017unsupervised} & M & \cmark & 0.182 &  1.481  & 6.501  & 0.267  & 0.725  & 0.906  & 0.963\\
Mahjourian~\cite{mahjourian2018unsupervised} & M & \cmark & 0.163 & 1.240 & 6.220 & 0.250 & 0.762 & 0.916 & 0.968\\
GeoNet~\cite{Yin18}\textdagger & M & \cmark   & 0.149 & 1.060 & 5.567 & 0.226 & 0.796 & 0.935 & 0.975\\
DDVO~\cite{wang2017learning} & M & \cmark & 0.151 & 1.257 & 5.583 & 0.228 & 0.810 & 0.936 & 0.974\\
DF-Net~\cite{zou2018df} & M & \cmark &  0.150 & 1.124 & 5.507 & 0.223 & 0.806 & 0.933 & 0.973\\
LEGO~\cite{yang2018lego} & M & \cmark &  0.162 & 1.352 & 6.276 & 0.252 & --- & --- & --- \\
Ranjan~\cite{ranjan2018adversarial}  & M & \cmark & 0.148 & 1.149 & 5.464 & 0.226 & 0.815 & 0.935 & 0.973\\
EPC++~\cite{luo2018every} & M & \cmark & 0.141 & 1.029 & 5.350 & 0.216 & 0.816 & 0.941 & 0.976\\
Struct2depth ~\cite{casser2018depth}  & M & \cmark & 0.141 & 1.026 & 5.291 &  0.215 & 0.816 & 0.945 & 0.979\\
Monodepth2\cite{monodepthv2} & M & \cmark &      0.115 &      \textbf{0.903} &      \textbf{4.863} &      \textbf{0.193 }&      \textbf{0.877} &      \textbf{0.959} &      \textbf{0.981} \\


\arrayrulecolor{grey}
\textit{Ours} & M & \xmark & \textbf{0.113} & 0.916 & 4.974 & 0.199 & 0.857 & 0.945 & 0.968 \\



\arrayrulecolor{black}\hline

\end{tabular}

\caption{Depth estimation accuracy on the KITTI test set. D  --- Depth supervision, D* --- Auxiliary depth supervision, M  --- Self-supervised mono, GT@Test --- uses elements of LiDAR ground truth at test time, \textdagger~--- Newer results from GitHub,+ pp --- With post-processing. For \textcolor{col1}{red} metrics, the lower is better; for \textcolor{col2}{blue} metrics, the higher is better. Best results in each category are in \textbf{bold}; second-best \underline{underlined}\vspace{-10pt}
    }
    \vspace{-10pt}

\label{tab:kitti_eigen}

\end{table}

%% file: tables/pseudo_road_len.tex
\begin{table}
  \small
  \centering

  \resizebox{0.9\textwidth}{!}{
  \begin{tabular}{|c|c|c|c|c|c|c|c|c|c|}
  \hline
  Length  & \cellcolor{col1}Abs Rel & \cellcolor{col1}Sq Rel & \cellcolor{col1}RMSE  & \cellcolor{col1}RMSE log & \cellcolor{col2}$\delta < 1.25 $ & \cellcolor{col2}$\delta < 1.25^{2}$ & \cellcolor{col2}$\delta < 1.25^{3}$\\
  
  \hline

6 & 0.338 & 2.857 & 7.47 & 0.353 & 0.068 & 0.132 & 0.155 \\
10 & 0.12 & 0.968 & 5.013 & 0.202 & 0.838 & 0.933 & 0.957 \\
15 & 0.116 & 0.942 & 5 & 0.2 & 0.853 & 0.944 & 0.968 \\
20 & 0.115 & 0.932 & 4.986 & 0.2 & 0.856 & 0.945 & 0.968 \\
25 & 0.117 & 0.956 & 5.002 & 0.201 & 0.856 & 0.944 & 0.969 \\
30 & 0.114 & 0.926 & 4.98 & 0.199 & 0.856 & 0.945 & 0.968 \\
40 & 0.116 & 0.933 & 4.985 & 0.2 & 0.856 & 0.946 & 0.971 \\
60 & 0.115 & 0.928 & 4.979 & 0.2 & 0.857 & 0.947 & 0.971 \\
80 & 0.116 & 0.941 & 4.99 & 0.2 & 0.857 & 0.945 & 0.97 \\

\arrayrulecolor{black}\hline

  \end{tabular}
  }
  \caption{Road model distance (length) ablation\vspace{-10pt}}
\label{tab:road_len} 
\vspace{-10pt}

\end{table}

%% file: tables/pseudo_road_width.tex
\begin{table}
  \centering
  \resizebox{0.9\textwidth}{!}{
  
  \begin{tabular}{|c|c|c|c|c|c|c|c|c|c|}
  \hline
  Width (m) & \cellcolor{col1}Abs Rel & \cellcolor{col1}Sq Rel & \cellcolor{col1}RMSE  & \cellcolor{col1}RMSE log & \cellcolor{col2}$\delta < 1.25 $ & \cellcolor{col2}$\delta < 1.25^{2}$ & \cellcolor{col2}$\delta < 1.25^{3}$\\
  
  \hline

0.5 & 0.119 & 0.944 & 5.02 & 0.206 & 0.838 & 0.932 & 0.957 \\
1 & 0.116 & 0.926 & 4.989 & 0.201 & 0.845 & 0.937 & 0.961 \\
2 & 0.114 & 0.918 & 4.981 & 0.2 & 0.856 & 0.944 & 0.968 \\
3 & 0.113 & 0.916 & 4.974 & 0.199 & 0.857 & 0.945 & 0.968 \\
4 & 0.115 & 0.936 & 4.99 & 0.2 & 0.856 & 0.945 & 0.968 \\
5 & 0.114 & 0.923 & 4.974 & 0.199 & 0.858 & 0.946 & 0.97 \\
10 & 0.116 & 0.933 & 4.986 & 0.2 & 0.855 & 0.946 & 0.969 \\
15 & 0.116 & 0.933 & 4.987 & 0.2 & 0.855 & 0.946 & 0.969 \\

\arrayrulecolor{black}\hline
  \end{tabular}
  }
    \caption{Road model width ablation. Points are considered to create the model if $|X| < \mbox{Width}$ and the distance is below 30 meters.\vspace{-10pt}}
\label{tab:road_width}  
\vspace{-10pt}
\end{table}

%% file: conclusion.tex
\section{Conclusion}
\label{sec:conclusion}
In this paper, we highlighted the limitation of self-supervised depth estimation methods and their reliance on LiDAR data at test time. We additionally showed how to overcome this issue by incorporating prior information about camera configuration and the environment, and we achieved comparable performance to the state of the art through vision only, without relying on any additional sensors.

%% file: main.bbl
\begin{thebibliography}{46}
\providecommand{\natexlab}[1]{#1}
\providecommand{\url}[1]{\texttt{#1}}
\expandafter\ifx\csname urlstyle\endcsname\relax
  \providecommand{\doi}[1]{doi: #1}\else
  \providecommand{\doi}{doi: \begingroup \urlstyle{rm}\Url}\fi

\bibitem[rad()]{radar-shortcomings}
Understanding radar for automotive ({ADAS}) solutions.
\newblock
  \url{https://www.pathpartnertech.com/understanding-radar-for-automotive-adas-solutions/}.
\newblock Accessed: 2019-11-20.

\bibitem[Bijelic et~al.(2018)Bijelic, Gruber, and Ritter]{Bijelic_2018}
Mario Bijelic, Tobias Gruber, and Werner Ritter.
\newblock A benchmark for lidar sensors in fog: Is detection breaking down?
\newblock In \emph{2018 IEEE Intelligent Vehicles Symposium (IV)}, pages
  760--767. IEEE, 2018.

\bibitem[Casser et~al.(2019{\natexlab{a}})Casser, Pirk, Mahjourian, and
  Angelova]{Casser19}
Vincent Casser, Soeren Pirk, Reza Mahjourian, and Anelia Angelova.
\newblock Unsupervised monocular depth and ego-motion learning with structure
  and semantics.
\newblock In \emph{Proceedings of the IEEE Conference on Computer Vision and
  Pattern Recognition Workshops}, pages 0--0, 2019{\natexlab{a}}.

\bibitem[Casser et~al.(2019{\natexlab{b}})Casser, Pirk, Mahjourian, and
  Angelova]{casser2018depth}
Vincent Casser, Soeren Pirk, Reza Mahjourian, and Anelia Angelova.
\newblock Depth prediction without the sensors: Leveraging structure for
  unsupervised learning from monocular videos.
\newblock In \emph{AAAI}, 2019{\natexlab{b}}.

\bibitem[Choe et~al.(2019)Choe, Joo, Rameau, Shim, and Kweon]{Segment2Regress}
Jaesung Choe, Kyungdon Joo, François Rameau, Gyu~Min Shim, and In~So Kweon.
\newblock Segment2regress: Monocular 3d vehicle localization in two stages.
\newblock In \emph{Robotics: Science and Systems}, 2019.

\bibitem[Eigen and Fergus(2015)]{EigenF14}
David Eigen and Rob Fergus.
\newblock Predicting depth, surface normals and semantic labels with a common
  multi-scale convolutional architecture.
\newblock In \emph{ICCV}, pages 2650--2658, 2015.

\bibitem[Faugeras et~al.(2001)Faugeras, Luong, and
  Papadopoulo]{faugeras01the-geometry}
Olivier~D. Faugeras, Quang{-}Tuan Luong, and Th{\'{e}}odore Papadopoulo.
\newblock \emph{The geometry of multiple images - the laws that govern the
  formation of multiple images of a scene and some of their applications}.
\newblock MIT Press, 2001.

\bibitem[Flynn et~al.(2016)Flynn, Neulander, Philbin, and Snavely]{FlynnNPS15}
John Flynn, Ivan Neulander, James Philbin, and Noah Snavely.
\newblock Deepstereo: Learning to predict new views from the world's imagery.
\newblock In \emph{ICCV}, pages 5515--5524, 2016.

\bibitem[Fu et~al.(2018)Fu, Gong, Wang, Batmanghelich, and Tao]{Fu_2018_CVPR}
Huan Fu, Mingming Gong, Chaohui Wang, Kayhan Batmanghelich, and Dacheng Tao.
\newblock Deep ordinal regression network for monocular depth estimation.
\newblock In \emph{CVPR}, June 2018.

\bibitem[Garg et~al.(2016)Garg, BG, Carneiro, and Reid]{GargBR16}
Ravi Garg, Vijay~Kumar BG, Gustavo Carneiro, and Ian Reid.
\newblock Unsupervised cnn for single view depth estimation: Geometry to the
  rescue.
\newblock In \emph{European Conference on Computer Vision}, pages 740--756.
  Springer, 2016.

\bibitem[Geiger et~al.(2012)Geiger, Lenz, and Urtasun]{Geiger2012CVPR}
Andreas Geiger, Philip Lenz, and Raquel Urtasun.
\newblock {Are we ready for Autonomous Driving? The KITTI Vision Benchmark
  Suite}.
\newblock In \emph{CVPR}, 2012.

\bibitem[Godard et~al.(2017)Godard, Mac~Aodha, and Brostow]{monodepth17}
Cl{\'e}ment Godard, Oisin Mac~Aodha, and Gabriel~J Brostow.
\newblock Unsupervised monocular depth estimation with left-right consistency.
\newblock In \emph{ICCV}, pages 270--279, 2017.

\bibitem[Godard et~al.(2019)Godard, Mac~Aodha, Firman, and
  Brostow]{monodepthv2}
Cl{\'e}ment Godard, Oisin Mac~Aodha, Michael Firman, and Gabriel~J Brostow.
\newblock Digging into self-supervised monocular depth estimation.
\newblock In \emph{Proceedings of the IEEE International Conference on Computer
  Vision}, pages 3828--3838, 2019.

\bibitem[Guizilini et~al.(2020)Guizilini, Ambrus, Pillai, Raventos, and
  Gaidon]{packnet}
Vitor Guizilini, Rares Ambrus, Sudeep Pillai, Allan Raventos, and Adrien
  Gaidon.
\newblock 3d packing for self-supervised monocular depth estimation.
\newblock In \emph{IEEE Conference on Computer Vision and Pattern Recognition
  (CVPR)}, 2020.

\bibitem[Guo et~al.(2018)Guo, Li, Yi, Ren, and Wang]{guo2018learning}
Xiaoyang Guo, Hongsheng Li, Shuai Yi, Jimmy Ren, and Xiaogang Wang.
\newblock Learning monocular depth by distilling cross-domain stereo networks.
\newblock In \emph{ECCV}, 2018.

\bibitem[Hartley and Zisserman(2004)]{Hartley2004}
R.~I. Hartley and A.~Zisserman.
\newblock \emph{Multiple View Geometry in Computer Vision}.
\newblock Cambridge University Press, ISBN: 0521540518, second edition, 2004.

\bibitem[Jaderberg et~al.(2015)Jaderberg, Simonyan, Zisserman,
  et~al.]{JaderbergSZK15}
Max Jaderberg, Karen Simonyan, Andrew Zisserman, et~al.
\newblock Spatial transformer networks.
\newblock In \emph{Advances in neural information processing systems}, pages
  2017--2025, 2015.

\bibitem[Klodt and Vedaldi(2018)]{klodt2018supervising}
Maria Klodt and Andrea Vedaldi.
\newblock Supervising the new with the old: learning {SFM} from {SFM}.
\newblock In \emph{ECCV}, 2018.

\bibitem[Kuznietsov et~al.(2017)Kuznietsov, Stuckler, and
  Leibe]{KuznietsovSL17}
Yevhen Kuznietsov, Jorg Stuckler, and Bastian Leibe.
\newblock Semi-supervised deep learning for monocular depth map prediction.
\newblock In \emph{ICCV}, pages 6647--6655, 2017.

\bibitem[Laina et~al.(2016)Laina, Rupprecht, Belagiannis, Tombari, and
  Navab]{LainaRBTN16}
Iro Laina, Christian Rupprecht, Vasileios Belagiannis, Federico Tombari, and
  Nassir Navab.
\newblock Deeper depth prediction with fully convolutional residual networks.
\newblock In \emph{2016 Fourth international conference on 3D vision (3DV)},
  pages 239--248. IEEE, 2016.

\bibitem[Lee et~al.(2019)Lee, Han, Ko, and Suh]{Lee19}
Jin~Han Lee, Myung-Kyu Han, Dong~Wook Ko, and Il~Hong Suh.
\newblock From big to small: Multi-scale local planar guidance for monocular
  depth estimation.
\newblock \emph{arXiv preprint arXiv:1907.10326}, 2019.

\bibitem[Liu et~al.(2015)Liu, Shen, Lin, and Reid]{LiuSLR15}
Fayao Liu, Chunhua Shen, Guosheng Lin, and Ian Reid.
\newblock Learning depth from single monocular images using deep convolutional
  neural fields.
\newblock \emph{IEEE TPAMI}, 38\penalty0 (10):\penalty0 2024--2039, 2015.

\bibitem[Luo et~al.(2018{\natexlab{a}})Luo, Yang, Wang, Wang, Xu, Nevatia, and
  Yuille]{luo2018every}
Chenxu Luo, Zhenheng Yang, Peng Wang, Yang Wang, Wei Xu, Ram Nevatia, and Alan
  Yuille.
\newblock Every pixel counts++: Joint learning of geometry and motion with {3D}
  holistic understanding.
\newblock \emph{arXiv}, 2018{\natexlab{a}}.

\bibitem[Luo et~al.(2018{\natexlab{b}})Luo, Ren, Lin, Pang, Sun, Li, and
  Lin]{singlestereo2018}
Yue Luo, Jimmy Ren, Mude Lin, Jiahao Pang, Wenxiu Sun, Hongsheng Li, and Liang
  Lin.
\newblock Single view stereo matching.
\newblock In \emph{CVPR}, 2018{\natexlab{b}}.

\bibitem[{Madhu Babu} et~al.(2018){Madhu Babu}, {Das}, {Majumdar}, and
  {Kumar}]{8593864}
V.~{Madhu Babu}, K.~{Das}, A.~{Majumdar}, and S.~{Kumar}.
\newblock Undemon: Unsupervised deep network for depth and ego-motion
  estimation.
\newblock In \emph{2018 IEEE/RSJ International Conference on Intelligent Robots
  and Systems (IROS)}, pages 1082--1088, Oct 2018.
\newblock \doi{10.1109/IROS.2018.8593864}.

\bibitem[Mahjourian et~al.(2018)Mahjourian, Wicke, and
  Angelova]{mahjourian2018unsupervised}
Reza Mahjourian, Martin Wicke, and Anelia Angelova.
\newblock Unsupervised learning of depth and ego-motion from monocular video
  using {3D} geometric constraints.
\newblock In \emph{CVPR}, 2018.

\bibitem[Man et~al.(2019)Man, Weng, Li, and Kitani]{groundnet}
Yunze Man, Xinshuo Weng, Xi~Li, and Kris Kitani.
\newblock Groundnet: Monocular ground plane normal estimation with geometric
  consistency.
\newblock In \emph{Proceedings of the 27th ACM International Conference on
  Multimedia}, MM ’19, page 2170–2178, New York, NY, USA, 2019. Association
  for Computing Machinery.
\newblock ISBN 9781450368896.
\newblock \doi{10.1145/3343031.3351068}.
\newblock URL \url{https://doi.org/10.1145/3343031.3351068}.

\bibitem[Mancini et~al.(2016)Mancini, Costante, Valigi, and
  Ciarfuglia]{mancini2016fast}
Michele Mancini, Gabriele Costante, Paolo Valigi, and Thomas~A Ciarfuglia.
\newblock Fast robust monocular depth estimation for obstacle detection with
  fully convolutional networks.
\newblock In \emph{2016 IEEE/RSJ International Conference on Intelligent Robots
  and Systems (IROS)}, pages 4296--4303. IEEE, 2016.

\bibitem[McCraith et~al.(2020)McCraith, Neumann, Zisserman, and
  Vedaldi]{mccraith2020monocular}
Robert McCraith, Lukas Neumann, Andrew Zisserman, and Andrea Vedaldi.
\newblock Monocular depth estimation with self-supervised instance adaptation,
  2020.

\bibitem[Nath~Kundu et~al.(2018)Nath~Kundu, Krishna~Uppala, Pahuja, and
  Babu]{gandepth2018}
Jogendra Nath~Kundu, Phani Krishna~Uppala, Anuj Pahuja, and R.~Venkatesh Babu.
\newblock {AdaDepth}: Unsupervised content congruent adaptation for depth
  estimation.
\newblock In \emph{CVPR}, 2018.

\bibitem[Poggi et~al.(2018)Poggi, Aleotti, Tosi, and
  Mattoccia]{poggi2018towards}
Matteo Poggi, Filippo Aleotti, Fabio Tosi, and Stefano Mattoccia.
\newblock Towards real-time unsupervised monocular depth estimation on cpu.
\newblock In \emph{2018 IEEE/RSJ International Conference on Intelligent Robots
  and Systems (IROS)}, pages 5848--5854. IEEE, 2018.

\bibitem[Ranjan et~al.(2019)Ranjan, Jampani, Kim, Sun, Wulff, and
  Black]{ranjan2018adversarial}
Anurag Ranjan, Varun Jampani, Kihwan Kim, Deqing Sun, Jonas Wulff, and
  Michael~J Black.
\newblock Competitive collaboration: Joint unsupervised learning of depth,
  camera motion, optical flow and motion segmentation.
\newblock In \emph{CVPR}, 2019.

\bibitem[Rousseeuw(1984)]{rousseeuw1984least}
Peter~J Rousseeuw.
\newblock Least median of squares regression.
\newblock \emph{Journal of the American statistical association}, 79\penalty0
  (388):\penalty0 871--880, 1984.

\bibitem[Scharstein et~al.(2002)Scharstein, Szeliski, and
  Szeliski]{scharstein2002a}
Daniel Scharstein, Richard Szeliski, and Rick Szeliski.
\newblock A taxonomy and evaluation of dense two-frame stereo correspondence
  algorithms.
\newblock \emph{International Journal of Computer Vision}, 47:\penalty0 7–42,
  May 2002.

\bibitem[Wang et~al.(2018)Wang, Buenaposada, Zhu, and Lucey]{wang2017learning}
Chaoyang Wang, Jose~Miguel Buenaposada, Rui Zhu, and Simon Lucey.
\newblock Learning depth from monocular videos using direct methods.
\newblock In \emph{CVPR}, 2018.

\bibitem[Wang et~al.(2004)Wang, Bovik, Sheikh, and Simoncelli]{wang2004image}
Zhou Wang, Alan~Conrad Bovik, Hamid~Rahim Sheikh, and Eero~P Simoncelli.
\newblock Image quality assessment: from error visibility to structural
  similarity.
\newblock \emph{TIP}, 2004.

\bibitem[Xie et~al.(2016)Xie, Girshick, and Farhadi]{XieGF16}
Junyuan Xie, Ross Girshick, and Ali Farhadi.
\newblock Deep3d: Fully automatic 2d-to-3d video conversion with deep
  convolutional neural networks.
\newblock In \emph{European Conference on Computer Vision}, pages 842--857.
  Springer, 2016.

\bibitem[Yang et~al.(2019)Yang, Hu, and Ramanan]{yang2019inferring}
Gengshan Yang, Peiyun Hu, and Deva Ramanan.
\newblock Inferring distributions over depth from a single image.
\newblock In \emph{2019 IEEE/RSJ International Conference on Intelligent Robots
  and Systems (IROS)}, 2019.

\bibitem[Yang et~al.(2018{\natexlab{a}})Yang, Wang, St{\"u}ckler, and
  Cremers]{yang2018deep}
Nan Yang, Rui Wang, J{\"o}rg St{\"u}ckler, and Daniel Cremers.
\newblock Deep virtual stereo odometry: Leveraging deep depth prediction for
  monocular direct sparse odometry.
\newblock In \emph{ECCV}, 2018{\natexlab{a}}.

\bibitem[Yang et~al.(2018{\natexlab{b}})Yang, Wang, Wang, Xu, and
  Nevatia]{yang2018lego}
Zhenheng Yang, Peng Wang, Yang Wang, Wei Xu, and Ram Nevatia.
\newblock {LEGO}: Learning edge with geometry all at once by watching videos.
\newblock In \emph{CVPR}, 2018{\natexlab{b}}.

\bibitem[Yang et~al.(2018{\natexlab{c}})Yang, Wang, Xu, Zhao, and
  Nevatia]{yang2017unsupervised}
Zhenheng Yang, Peng Wang, Wei Xu, Liang Zhao, and Ramakant Nevatia.
\newblock Unsupervised learning of geometry with edge-aware depth-normal
  consistency.
\newblock In \emph{AAAI}, 2018{\natexlab{c}}.

\bibitem[Yin and Shi(2018)]{Yin18}
Zhichao Yin and Jianping Shi.
\newblock Geonet: Unsupervised learning of dense depth, optical flow and camera
  pose.
\newblock In \emph{Proceedings of the IEEE Conference on Computer Vision and
  Pattern Recognition}, pages 1983--1992, 2018.

\bibitem[Zhao et~al.(2016)Zhao, Gallo, Frosio, and Kautz]{lossfunctions}
Hang Zhao, Orazio Gallo, Iuri Frosio, and Jan Kautz.
\newblock Loss functions for image restoration with neural networks.
\newblock \emph{IEEE Transactions on computational imaging}, 3\penalty0
  (1):\penalty0 47--57, 2016.

\bibitem[Zhou et~al.(2018)Zhou, Zhao, Puig, Xiao, Fidler, Barriuso, and
  Torralba]{zhou2018semantic}
Bolei Zhou, Hang Zhao, Xavier Puig, Tete Xiao, Sanja Fidler, Adela Barriuso,
  and Antonio Torralba.
\newblock Semantic understanding of scenes through the ade20k dataset.
\newblock \emph{International Journal on Computer Vision}, 2018.

\bibitem[Zhou et~al.(2017)Zhou, Brown, Snavely, and Lowe]{zhou2017}
Tinghui Zhou, Matthew Brown, Noah Snavely, and David~G Lowe.
\newblock Unsupervised learning of depth and ego-motion from video.
\newblock In \emph{Proceedings of the IEEE Conference on Computer Vision and
  Pattern Recognition}, pages 1851--1858, 2017.

\bibitem[Zou et~al.(2018)Zou, Luo, and Huang]{zou2018df}
Yuliang Zou, Zelun Luo, and Jia-Bin Huang.
\newblock {DF-Net}: Unsupervised joint learning of depth and flow using
  cross-task consistency.
\newblock In \emph{ECCV}, 2018.

\end{thebibliography}
